\documentclass[sigconf]{acmart}

\usepackage{dblfloatfix}

\def\BibTeX{{\rm B\kern-.05em{\sc i\kern-.025em b}\kern-.08emT\kern-.1667em\lower.7ex\hbox{E}\kern-.125emX}}

\copyrightyear{2019}
\acmYear{2019}
\setcopyright{rightsretained}
\acmConference[IWOCL'19]{International Workshop on OpenCL}{May 13--15, 2019}{Boston, MA, USA}
\acmBooktitle{International Workshop on OpenCL (IWOCL'19), May 13--15, 2019, Boston, MA, USA}
\acmDOI{10.1145/3318170.3318183}
\acmISBN{978-1-4503-6230-6/19/05}

\begin{document}

%
\title{Accelerated Neural Networks on OpenCL Devices Using SYCL-DNN}

%
\author{Rod Burns, John Lawson, Duncan McBain and Daniel Soutar}
\authornote{Authors listed alphabetically}
\email{{rod, john, duncan, daniel.soutar}@codeplay.com}
\affiliation{%
  \institution{Codeplay Software Ltd.}
  \streetaddress{Level C, Argyle House, 3 Lady Lawson Street}
  \city{Edinburgh}
  \country{UK}
  \postcode{EH3 9DR}
}

%
\renewcommand{\shortauthors}{Burns, Lawson, McBain and Soutar}

%
\begin{abstract}

  Over the past few years machine learning has seen a renewed explosion of
  interest, following a number of studies showing the effectiveness of neural
  networks in a range of tasks which had previously been considered incredibly
  hard. Neural networks' effectiveness in the fields of image recognition and
  natural language processing stems primarily from the vast amounts of data
  available to companies and researchers, coupled with the huge amounts of
  compute power available in modern accelerators such as GPUs, FPGAs and ASICs.
  There are a number of approaches available to developers for utilizing GPGPU
  technologies such as SYCL, OpenCL and CUDA, however many applications require
  the same low level mathematical routines. Libraries dedicated to accelerating
  these common routines allow developers to easily make full use of the
  available hardware without requiring low level knowledge of the hardware
  themselves, however such libraries are often provided by hardware
  manufacturers for specific hardware such as cuDNN~\cite{cudnn} for Nvidia
  hardware or MIOpen~\cite{miopen} for AMD hardware.

  SYCL-DNN is a new open-source library dedicated to providing accelerated
  routines for neural network operations which are hardware and vendor agnostic.
  Built on top of the SYCL open standard and written entirely in standard C++,
  SYCL-DNN allows a user to easily accelerate neural network code for a wide
  range of hardware using a modern C++ interface. The library is tested on AMD's
  OpenCL for GPU, Intel's OpenCL for CPU and GPU, ARM's OpenCL for Mali GPUs as
  well as ComputeAorta's OpenCL for R-Car CV engine and host CPU\@. In this talk
  we will present performance figures for SYCL-DNN on this range of hardware,
  and discuss how high performance was achieved on such a varied set of
  accelerators with such different hardware features.

\end{abstract}

%
%


%

%

%
\maketitle

\section{Introduction}

Deep neural networks (DNNs) have been widely studied in the past few years, as
they have repeatedly proved effective at solving hard computational problems
when trained on sufficiently large data sets.

The resurgence in study of DNNs primarily started in 2012 when
AlexNet~\cite{AlexNet}, a convolutional neural network (CNN), beat all previous
entries to the ImageNet competition~\cite{ILSVRC15} with an error rate of 15.3\%
on the Top-1 task, compared to that year's next best 26.2\%. Since then the
error rates shown in the competition results have plummeted with all entries
using DNNs.

The effectiveness of these neural networks has been achieved through both the
large datasets now available and the vast amounts of compute provided by GPU
hardware and other hardware acceleration. While CPUs offer a few highly
programmable and fast cores, GPUs provide many more cores which are more
limited. This restricts the tasks which can be effectively accelerated on GPU
hardware, but for many of the numeric tasks in DNNs this many core, highly
parallel model fits well. These common tasks include routines such as matrix
multiplies and convolutions which typically make up the majority of the runtime
of a neural network. As such any performance improvements obtained in these
routines give substantial improvements to the performance of the whole DNN\@.

Massively parallel hardware is well suited to these tasks yet requires a very
different application model and low level knowledge of the hardware to achieve
good performance. This can be a barrier to obtaining good performance in machine
learning models so many hardware vendors provide libraries targeting their
platforms to accelerate these standard routines, such as NVidia's
cuDNN~\cite{cudnn} and AMD's MIOpen~\cite{miopen}. These libraries are typically
tuned for the specific hardware they run on and are implemented in low level
assembly or some intermediate representation to get optimal performance.

\begin{figure*}[!tbp]
  \includegraphics{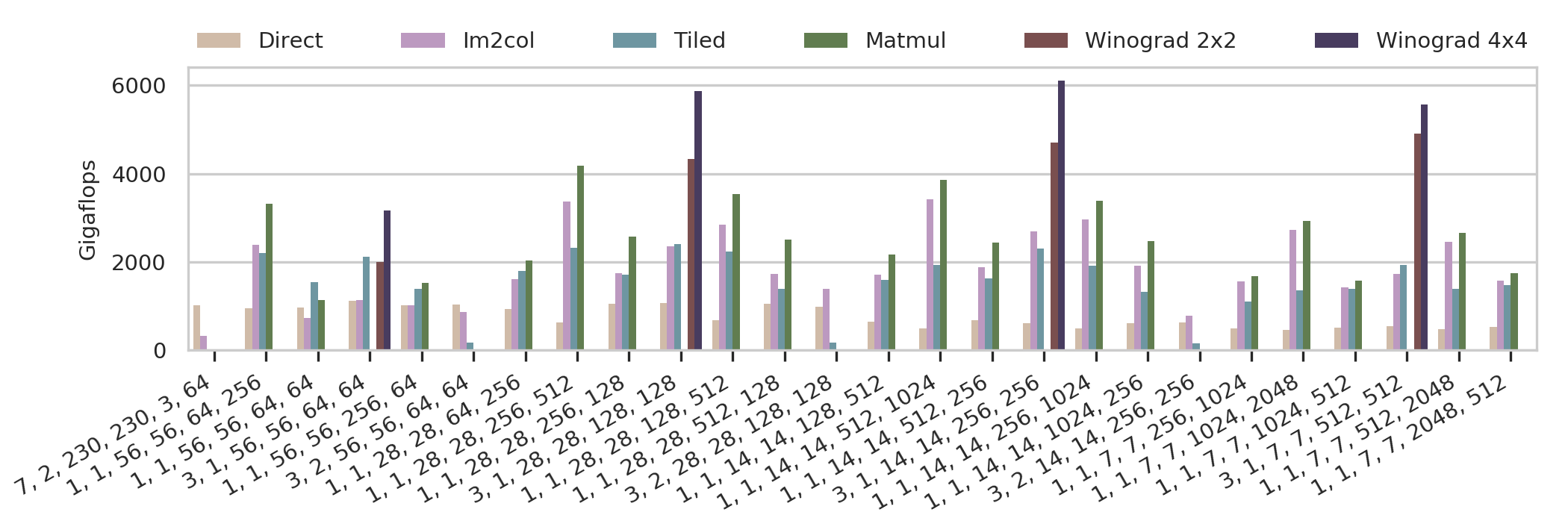}%
  \Description{A bar chart comparing the performance of 6 six different
  algorithms over 26 sets of convolutions. The algorithms perform differently
  for different sets of parameters, with the tiled algorithm often performing
  best for convolutions with a window size of 1 achieving between 1.8 and 4
  teraflops, while the Winograd technique often performs best for 3 by 3
  windows, achieving up to 6 teraflops.
  }%
  \caption{The number of gigaflops achieved by different algorithms for a range of
    convolutions from ResNet-50 with a batch size of 32 on an AMD R9 Nano. The
    convolution parameters are given by: window size, stride, image rows, image
    columns, input features, output features. Not all algorithms are compatible
    with all sets of parameters.
  }%
  \label{fig:amd_selectors}%
\end{figure*}

\begin{figure*}
  \includegraphics{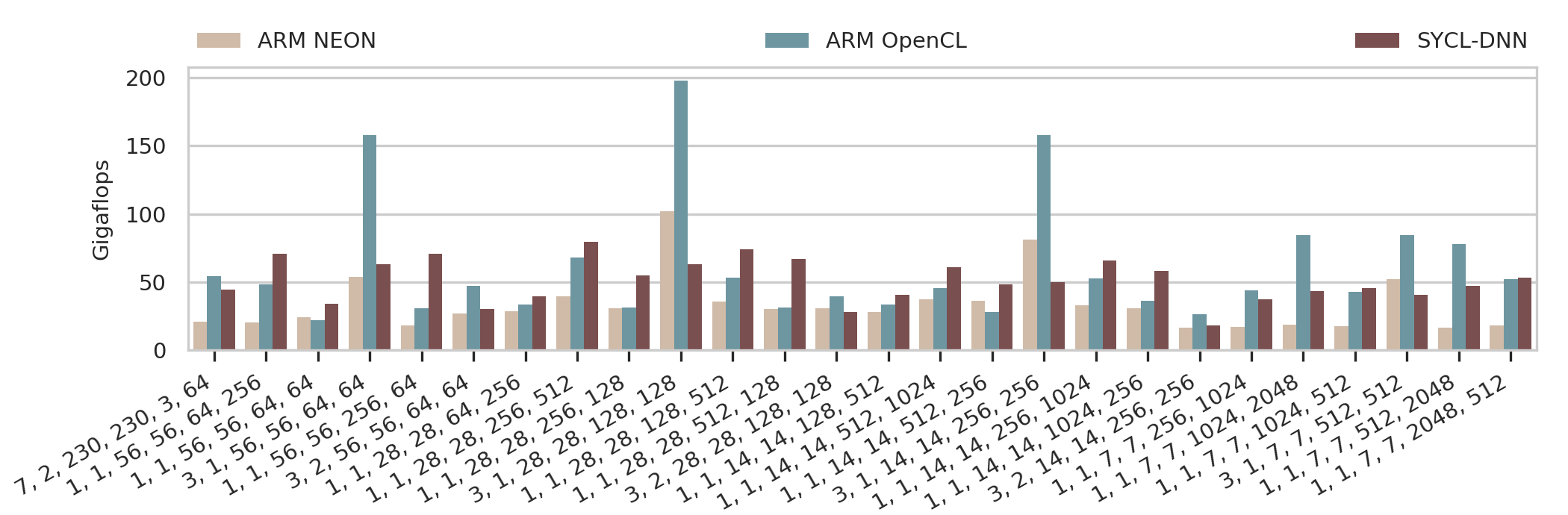}%
  \Description{A bar chart comparing SYCL-DNN to ARM Compute Library using NEON
  and OpenCL across 26 sets of convolutions. For most of the convolution sizes
  SYCL-DNN performs better than ARM Compute Library, and their OpenCL
  implementation is always faster than their NEON implementation. SYCL-DNN
  typically gets between 40 and 80 gigaflops, while in most cases NEON reaches a
  maximum of around 30 and OpenCL around 100. There are three convolutions which
  have a window size of 3, where ARM Compute Library is significantly faster,
  achieving over 150 gigaflops with their OpenCL implementation, while in these
  cases SYCL-DNN only achieves around 60.
  }%
  \caption{The number of gigaflops achieved on the ARM HiKey 960 SoC, with
  SYCL-DNN running on the Mali G-71 GPU compared to ARM's Compute Library
  running on the Mali G-71 GPU using OpenCL and on the CPU using NEON\@. The
  graph shows the convolutions in ResNet-50, run with a batch size of 1.
  }%
  \label{fig:arm_resnet}
\end{figure*}

\begin{figure*}
  \includegraphics{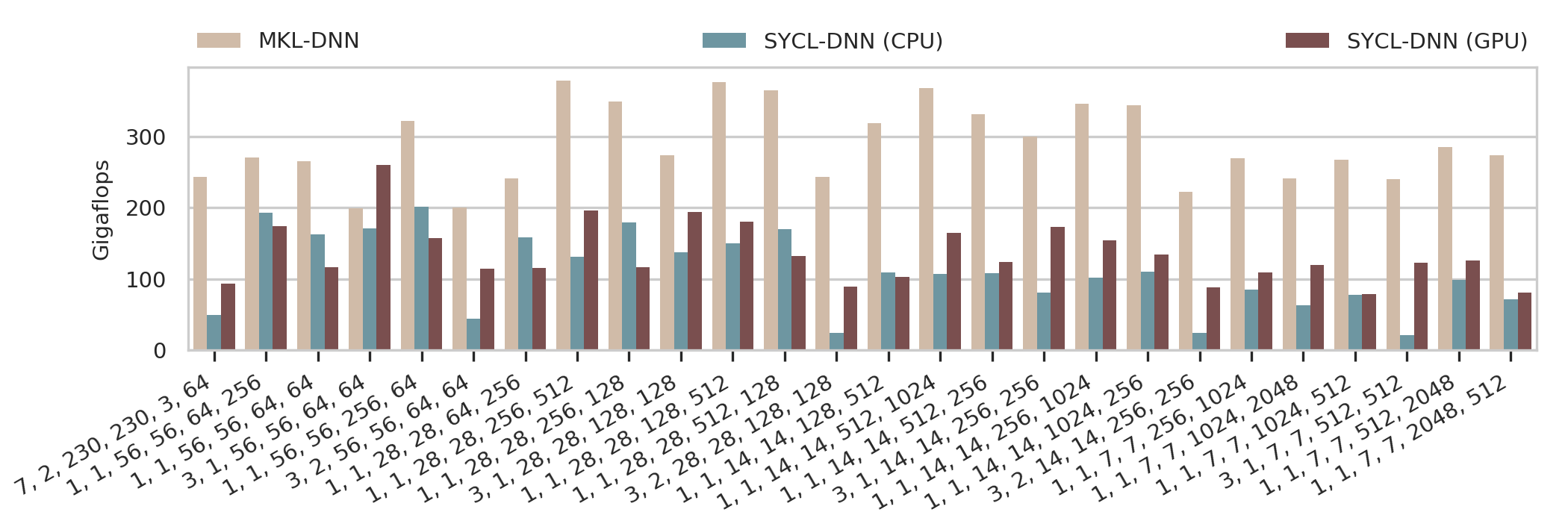}
  \Description{A bar chart comparing SYCL-DNN to MKL-DNN across 26 different
  convolution sizes. MKL-DNN is faster than SYCL-DNN in almost all cases,
  typically achieving 200 to 380 gigaflops while SYCL-DNN using either the GPU
  or CPU achieves between 100 and 200. Most of the time the GPU SYCL-DNN
  implementation is faster than the CPU SYCL-DNN implementation, however there
  are a few cases where it is not.
  }%
  \caption{The number of gigaflops achieved on the Intel i7-6700K processor, with
  SYCL-DNN running on the integrated GPU and on the CPU compared to MKL-DNN\@.
  The graph shows the convolutions in ResNet-50, run with a batch size of 4.}%
  \label{fig:intel_resnet}
\end{figure*}

\section{The SYCL programming model}

SYCL~\cite{sycl} is a royalty-free open standard maintained by the Khronos group
which provides a high level abstraction of GPGPU concepts based on
OpenCL~\cite{opencl}. Using SYCL a developer can write standard C++ code to be
run on accelerators supporting OpenCL, and so use the functionality
provided by modern C++ including templates, inheritance, metaprogramming and the
standard library while at the same time benefiting from the capabilities of the
underlying hardware as exposed through OpenCL\@.

The design of SYCL is built around a single source programming model in
completely standard C++, in contrast to other similar programming models which
typically require additional keywords and restrictions on the language. By
building on top of standard C++, the SYCL standard inherits all the improvements
made recently to the language along with support from many standard tools.

Underneath, SYCL interacts with OpenCL devices and so hardware manufacturers do
not need to provide any further implementation to allow developers to use their
devices, provided a compatible OpenCL driver is available.

As a royalty-free open standard, SYCL can be implemented by anyone. At the time of
writing the only fully conformant SYCL implementation is Codeplay Software's
ComputeCpp~\cite{computecpp}, which we use in all the following benchmarks.
Other implementations available include triSYCL~\cite{trisycl} and
hip-SYCL~\cite{hipsycl}.

\section{SYCL-DNN}

SYCL-DNN~\cite{sycldnn} is an open-source library developed by Codeplay Software
to accelerate machine learning applications on OpenCL hardware using the SYCL
programming model. The library has been specifically tuned for certain devices,
but is usable on any device supported by the user's choice of SYCL
implementation.

The library provides a high level interface allowing users to run neural network
primitive routines accelerated on OpenCL hardware. Internally it contains a
number of highly parameterized SYCL kernels to provide the computations required
for each routine, and for some routines it provides a variety of different
algorithms which all provide the same numeric results. In this way the library
can adapt to different hardware characteristics by either choosing different
algorithms or different parameters for each algorithm to maximize performance on
the hardware.

Such customization is currently provided through manual tuning for a target
device, however an automated, intelligent approach is planned for the future.

Convolutions are the main component of modern image recognition networks,
providing a mechanism to detect features in images which is invariant with
respect to the feature's position in the image. State-of-the-art image
recognition networks such as VGG~\cite{vgg} and ResNet~\cite{resnet} are made up
of many layers of convolutions, interspersed with pooling and normalization
layers. The convolutions are the most compute intensive operations in these
networks and so the primary optimization target in SYCL-DNN\@.

SYCL-DNN provides a number of implementations of a convolution, from a
vectorized naive compute kernel which runs a single thread per output vector, to
a tiled Winograd operation~\cite{winograd} which uses data transforms to convert
the convolution into a number of small matrix multiplies, reducing
the total number of floating point operations.

Figure~\ref{fig:amd_selectors} shows the performance of the different
convolution algorithms implemented in SYCL-DNN for a range of different
convolutions in the ResNet-50 DNN model~\cite{resnet}. For different
convolution parameters different algorithms perform better than others, with no
single algorithm always performing best.

\section{Performance across devices}

The number of devices that can support SYCL-DNN is large, as SYCL-DNN is written
in SYCL which can run on many OpenCL implementations. The primary
target of the library has been embedded devices, though with few changes the
same code can target desktop GPUs and other hardware.

The embedded devices that the library has been targeted towards are ARM's Mali
G-71 GPU---a high-end mobile GPU that is designed for low power devices---and
Renesas' R-Car platforms like the V3H---a system on chip designed for automated
driving solutions that provide a programmable CV engine accelerator.

In addition to these, the library is tested on Intel processors, making use of
both their CPU OpenCL implementation and the Intel Compute Runtime that targets
integrated GPUs, and on AMD GPUs with both Fiji and Hawaii devices tested.

Figures~\ref{fig:amd_selectors},~\ref{fig:arm_resnet} and~\ref{fig:intel_resnet}
show the performance achieved using SYCL-DNN on a range of hardware when running
the convolutional layers from ResNet-50.

The performance results for ResNet convolutions on an Intel i7-6700K processor
are shown in Figure~\ref{fig:intel_resnet}, where SYCL-DNN can use either the
CPU or the GPU\@. The performance is compared against
MKL-DNN~\cite[v0.18.1]{mkl_dnn}, which contains highly vectorized micro-kernels
that are JIT compiled to best suit particular tasks. Compared to other
platforms, less work has been channeled into optimizing for Intel processors
which is evident when compared to the very good performance that MKL-DNN
achieves.

On the HiKey 960 system-on-chip (SoC), SYCL-DNN utilises the ARM Mali G-71 GPU and
consistently outperforms ARM's Compute Library~\cite[v18.11]{arm_compute} which
runs on both the CPU using NEON and on the GPU using OpenCL, as shown in
Figure~\ref{fig:arm_resnet}. Three of the convolution parameters stand out where
the Compute Library achieves over 150 gigaflops; these are 3$\times$3
convolutions without a stride which are very well optimized in their library. In
almost all other cases SYCL-DNN achieves better performance than the hand tuned
OpenCL kernels in the Compute Library.

\section{Conclusion}

Overall the performance provided by our library is good, but there are still
significant improvements to be made. Further optimizations that are planned
include implementing better support for memory prefetching and better
utilization of user programmable caches, as well as providing alternative
implementations of the compute routines which may perform better on some
hardware platforms.

The SYCL-DNN team intend to extend the devices that the library is tuned for,
and develop additional automated approaches to further ease this process in the
future. As these tuning decisions involve many parameters and a large number of
features they are a good candidate for a learned solution rather than a hand
tuned one.

%
\bibliographystyle{ACM-Reference-Format}
\bibliography{iwocl2019-sycldnn}

\end{document}